\documentclass[conference]{IEEEtran}
\usepackage{cite}
\usepackage{amsmath,amssymb,amsfonts}
\usepackage{algorithmic}
\usepackage{graphicx}
\graphicspath{{./}}
\usepackage{textcomp}
\usepackage{xcolor}
\usepackage{booktabs}
\usepackage{subcaption}

\def\BibTeX{{\rm B\kern-.05em{\sc i\kern-.025em b}\kern-.08em
        T\kern-.1667em\lower.7ex\hbox{E}\kern-.125emX}}

\begin{document}
	\setlength{\columnsep}{0.22in}

    \title{SpaceRipple: Lightweight Semantic Delivery for Mission-Oriented LEO Earth Observation Satellite Networks}

    \author{
        \IEEEauthorblockN{Ziyi Yang, Hao Yuan, Yunxiang Yi, Wenbo Wang and Xing Zhang}
        \IEEEauthorblockA{
            \textit{Beijing University of Posts and Telecommunications} \\
            Beijing, China
        }
    }

    \maketitle

    \begin{abstract}
        Earth observation satellite networks generate massive volumes of high-resolution imagery, whereas inter-satellite and downlink resources remain limited. In many time-sensitive missions, ground users require mission-relevant semantic information rather than a full raw-image downlink. This paper proposes SpaceRipple, a lightweight framework for mission-oriented semantic delivery and on-board processing in Earth observation satellite networks. A sensing satellite performs adaptive compression and metadata generation to reduce inter-satellite traffic, while an edge computing satellite restores the received representation and extracts task-relevant semantic information. Unlike fidelity-driven image transmission, SpaceRipple coordinates compression, forwarding, restoration, and semantic inference within a collaborative pipeline, enabling semantic-oriented delivery instead of pixel-level image delivery. A compression-aware MoE enhancement module is further introduced to improve robustness under degraded visual inputs. Experimental results show that SpaceRipple achieves favorable reconstruction quality, improved semantic detection performance, and substantial bandwidth savings, demonstrating its potential for efficient and reliable Earth observation under constrained satellite-network resources.
    \end{abstract}

    \begin{IEEEkeywords}
        Earth observation satellite networks, on-board image processing, communication-computing co-design, semantic communication, mixture of experts
    \end{IEEEkeywords}

    \section{Introduction}

    Earth observation satellites support disaster monitoring, target recognition, environmental assessment, and resource investigation \cite{wang2025access}. However, increasing image resolution sharply raises transmission cost. The conventional ``acquisition--full downlink--ground processing'' pipeline is therefore inefficient for time-sensitive missions \cite{chen2024spaceedge,sun2025deepspace}.

    Advances in inter-satellite links and on-board computing make collaborative processing increasingly feasible \cite{wang2025access,chen2024spaceedge}. In many scenarios, ground operators do not require full raw imagery; instead, they need task-relevant semantic information such as scene labels, detected targets, or regional status indicators. This motivates semantic-oriented delivery rather than fidelity-driven image downlink \cite{xin2024semsurvey,guo2024kbisc}. However, requiring the sensing satellite to generate only compact semantic messages may reduce adaptability across heterogeneous missions and weaken robustness under varying observation quality. A practical system should therefore preserve sufficient intermediate information for downstream recovery and task-oriented refinement, while still avoiding full-image downlink.

    Existing studies have addressed image compression, on-board intelligence, and semantic communication separately \cite{wang2022taskcompression,wang2025access,sun2025deepspace}. Recent SR-powered satellite image acquisition systems show that on-board compression and receiver-side reconstruction can substantially reduce downlink traffic, but they still primarily target high-fidelity image acquisition or reconstruction \cite{sun2025deepspace}. SpaceRipple instead treats restoration as an intermediate step toward semantic-oriented delivery, where the final downlink objective is mission-relevant semantic information rather than reconstructed imagery. The sensing satellite must reduce payload size while preserving task-relevant cues, and the edge computing satellite must recover a representation suitable for reliable semantic inference.

    We therefore propose SpaceRipple, a lightweight and reliable framework for mission-oriented semantic delivery and on-board processing in Earth observation satellite networks. The sensing satellite performs adaptive compression and metadata generation, the edge computing satellite conducts restoration, enhancement, and semantic inference, and the ground receives compact semantic messages instead of full image downlink. The novelty of SpaceRipple lies primarily in the system-level co-design of compression, forwarding, restoration, and semantic-oriented delivery for mission-oriented Earth observation satellite networks. An MoE-based enhancement module is further introduced to improve robustness after compression \cite{riquelme2021vmoe,sun2025deepspace}.

    The main contributions are as follows:
    \begin{itemize}
        \item We present a collaborative satellite-network framework that couples image compression, inter-satellite forwarding, restoration, and semantic-oriented delivery for Earth observation.
        \item We introduce a metadata-assisted restoration pipeline with pixel-feature-driven MoE enhancement to improve task robustness under compressed transmission.
        \item We evaluate the framework from visual, task, and system perspectives, showing strong reconstruction quality, higher semantic F1 scores, and clear contact-window throughput gains.
    \end{itemize}

    \section{System Model}

    \subsection{System Architecture}

    \begin{figure}[!t]
        \centering
        \includegraphics[width=0.9\columnwidth]{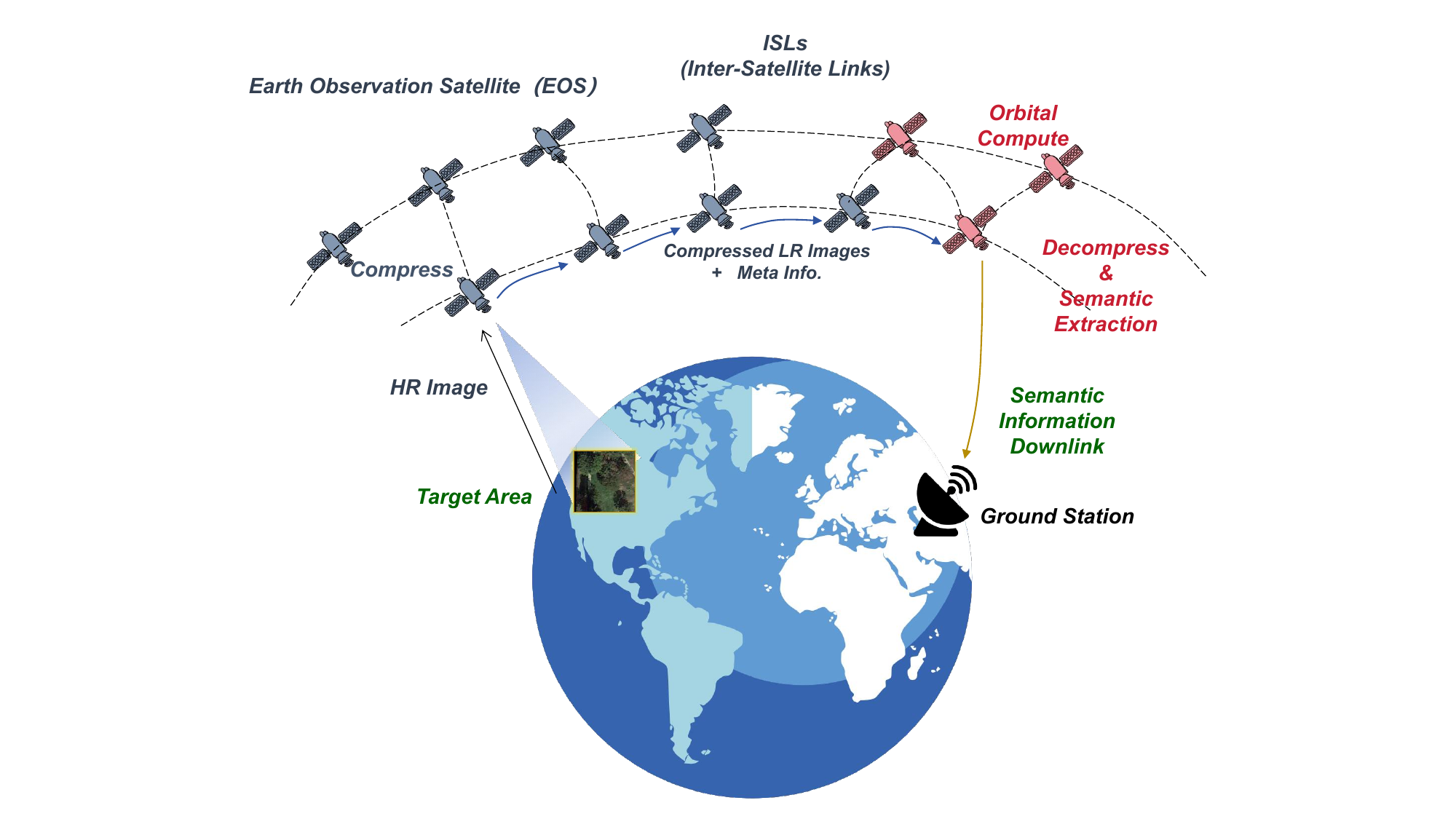}
        \caption{Application scenario of SpaceRipple.}
        \label{fig:scenario_overview}
    \end{figure}

    \begin{figure*}[!t]
        \centering
        \includegraphics[width=0.8\textwidth]{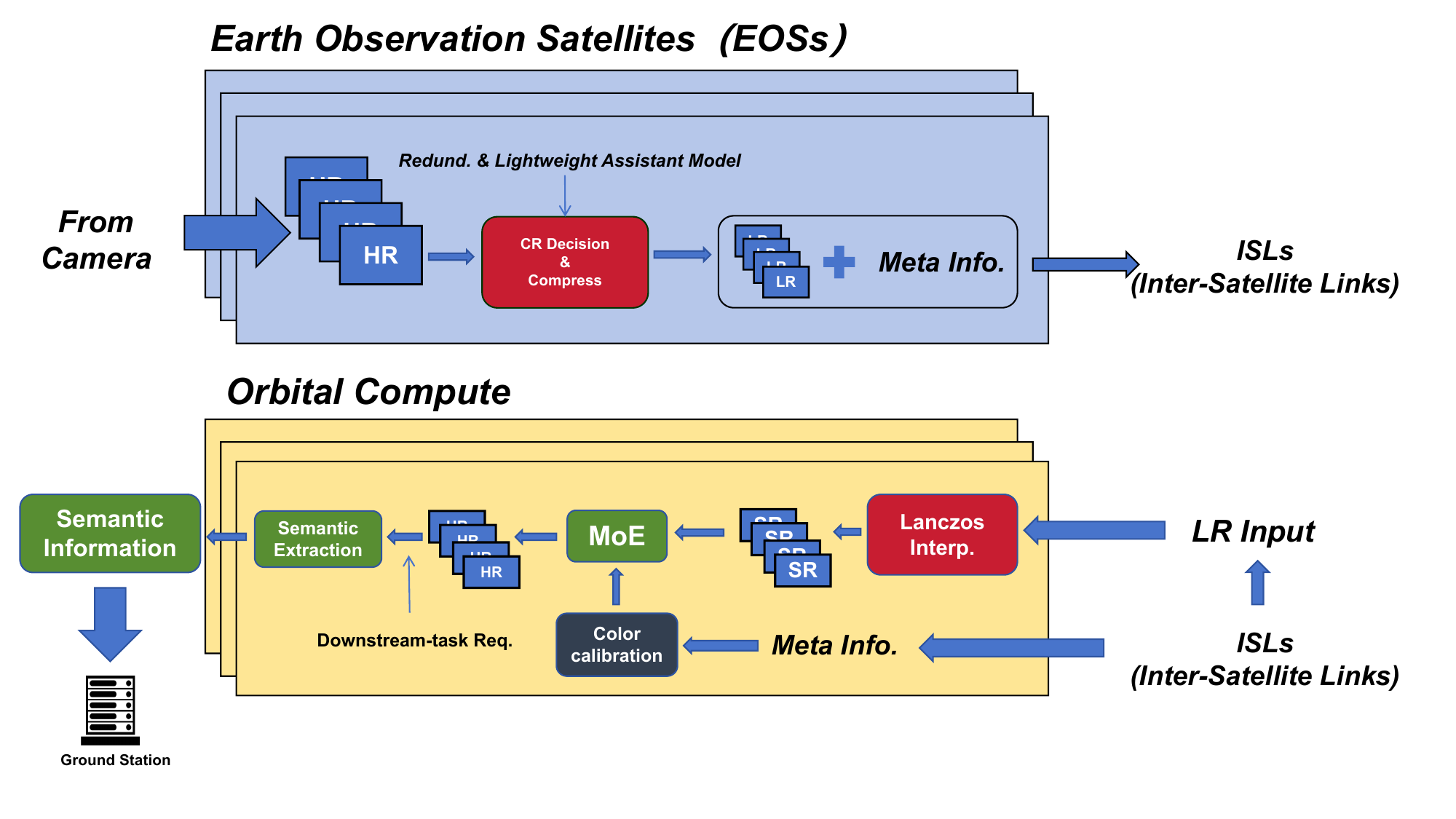}
        \caption{System workflow of SpaceRipple.}
        \label{fig:system_overview}
    \end{figure*}

    SpaceRipple comprises four functional components: sensing satellites, inter-satellite links, edge computing satellites, and ground stations. As shown in Fig.~\ref{fig:scenario_overview}, the sensing satellite acquires high-resolution imagery, performs on-board compression, and forwards the compressed image together with metadata to an edge computing satellite. The edge computing satellite then performs restoration and semantic extraction, while the ground station receives mission-relevant semantic messages instead of a full-image downlink. Fig.~\ref{fig:system_overview} illustrates the workflow. This architecture reduces the sensing-side burden and avoids raw-image downlink, improving mission-oriented information delivery under constrained transmission and on-board resources.

    \subsection{Data Flow and Processing Pipeline}

    According to the system workflow, SpaceRipple consists of four stages: observation, compression and forwarding, on-board restoration and semantic processing, and semantic-result downlink. The sensing satellite captures a raw high-resolution image and compresses it into a communication-efficient representation together with metadata. The edge computing satellite then restores and enhances the received input before semantic inference \cite{wang2022taskcompression,guo2024kbisc}.

    In the final stage, only the semantic message is downlinked to the ground, rather than the complete raw image or a fidelity-optimized reconstruction. Accordingly, the downlink objective shifts from pixel-level image delivery to mission-relevant information delivery, shortening the ``observation--transmission--decision'' cycle.

    \subsection{Data Representation and Modeling}

    Let the raw image captured by the sensing satellite be denoted by $I_{HR}$, where the subscript $HR$ indicates the original high-resolution observation. The sensing-side compression module $f_{\phi}(\cdot)$ maps the raw image to a compressed representation and associated side metadata:
    \begin{equation}
        (I_C,M)=f_{\phi}(I_{HR}).
        \label{eq:compress_pipeline}
    \end{equation}
    Here, $I_C$ denotes the compressed image representation to be transmitted, and $M$ denotes the side metadata generated during compression.

    In SpaceRipple, the metadata is represented as
    \begin{equation}
        M=\{r,c\},
        \label{eq:metadata}
    \end{equation}
    where $r$ denotes the compression ratio and $c$ denotes calibration-related side information, such as color calibration cues. The transmitted packet is therefore written as
    \begin{equation}
        X=\{I_C,M\},
        \label{eq:packet}
    \end{equation}
    where $X$ denotes the complete transmitted representation delivered through the inter-satellite link. Its payload size is
    \begin{equation}
        B_{\text{pkt}}=B(I_C)+B(M),
        \label{eq:payload}
    \end{equation}
    where $B(\cdot)$ denotes the number of transmitted bits.

    Upon reception, the edge computing satellite performs restoration as
    \begin{equation}
        I_R=g_{\psi}(I_C,M),
        \label{eq:restore}
    \end{equation}
    where $g_{\psi}(\cdot)$ denotes the restoration module and $I_R$ denotes the restored image used for downstream semantic inference. To characterize structural recoverability after compression, we further define a recoverability score based on the structural similarity between the restored image and the original high-resolution image:
    \begin{equation}
        R_{\mathrm{red}}=\mathrm{SSIM}(I_R,I_{HR}),
        \label{eq:redundancy_ssim}
    \end{equation}
    where $\mathrm{SSIM}(\cdot,\cdot)$ denotes the structural similarity index. A higher value of $R_{\mathrm{red}}$ indicates that more structural information from the original image remains recoverable after compression and restoration.

    Finally, the semantic head $h_{\omega}(\cdot)$ produces the mission-oriented output
    \begin{equation}
        S=h_{\omega}(I_R),
        \label{eq:semantic}
    \end{equation}
    where $h_{\omega}(\cdot)$ denotes the semantic inference head and $S$ denotes the final semantic output, which may correspond to scene labels, detection results, or regional attributes depending on the specific task.

    The objective is not exact pixel-level reconstruction, but the reliable preservation of task-relevant information needed to support mission inference under constrained transmission resources \cite{guo2024kbisc}.

    \subsection{Design Objectives and Characteristics}

    The design of SpaceRipple is guided by transmission efficiency, processing reliability, and communication--computing co-design.

    First, the framework reduces transmission burden. By performing adaptive compression on the sensing satellite and shifting restoration and semantic processing to the edge computing satellite, SpaceRipple avoids forwarding raw imagery and uses limited link resources more efficiently \cite{wang2022taskcompression}.

    Second, efficiency should not undermine mission robustness. SpaceRipple therefore transmits metadata together with the compressed image and performs restoration, calibration, and enhancement on the edge computing satellite, helping compensate for compression-induced loss and stabilize semantic outputs \cite{guo2024kbisc}.

    At the system level, the design preference of SpaceRipple can be summarized as
    \begin{equation}
        \mathcal{J}_{\text{sys}} \propto Q_{\text{task}} \uparrow,\quad B_{\text{pkt}} \downarrow,\quad P_{\text{obs}} \downarrow,
        \label{eq:jsys}
    \end{equation}
    where $\mathcal{J}_{\text{sys}}$ denotes the overall system utility, $Q_{\text{task}}$ denotes mission-task effectiveness, $B_{\text{pkt}}$ is the transmitted payload size per sample, and $P_{\text{obs}}$ is the sensing-side model footprint. This relation expresses the intended design tradeoff rather than an explicitly optimized training objective.

    \section{Proposed Method}

    SpaceRipple follows a task-oriented design principle: imagery is compressed at the sensing satellite and semantically refined at the edge computing satellite. Rather than optimizing image fidelity alone, the framework jointly considers transmission efficiency, restoration usability, and downstream semantic effectiveness under constrained inter-satellite resources \cite{guo2024kbisc}. The proposed method is designed as a system-level collaborative pipeline rather than as a standalone restoration backbone.

    \subsection{Adaptive Compression, Metadata, and Restoration}

    Given a raw image, the sensing satellite first performs adaptive compression to generate a communication-efficient representation together with metadata \cite{wang2022taskcompression,sun2025deepspace}. According to \eqref{eq:metadata}--\eqref{eq:payload}, the transmitted information consists of the compressed image and metadata carrying compression-ratio and calibration-related cues. The compression ratio is defined as
    \begin{equation}
    	CR = \frac{\text{HR size}}{\text{LR size}},
    \end{equation}
    where \textit{HR size} and \textit{LR size} denote the source-image size and the compressed payload size, respectively.
    \begin{figure*}[!t]
    	\centering
    	\includegraphics[width=0.88\textwidth]{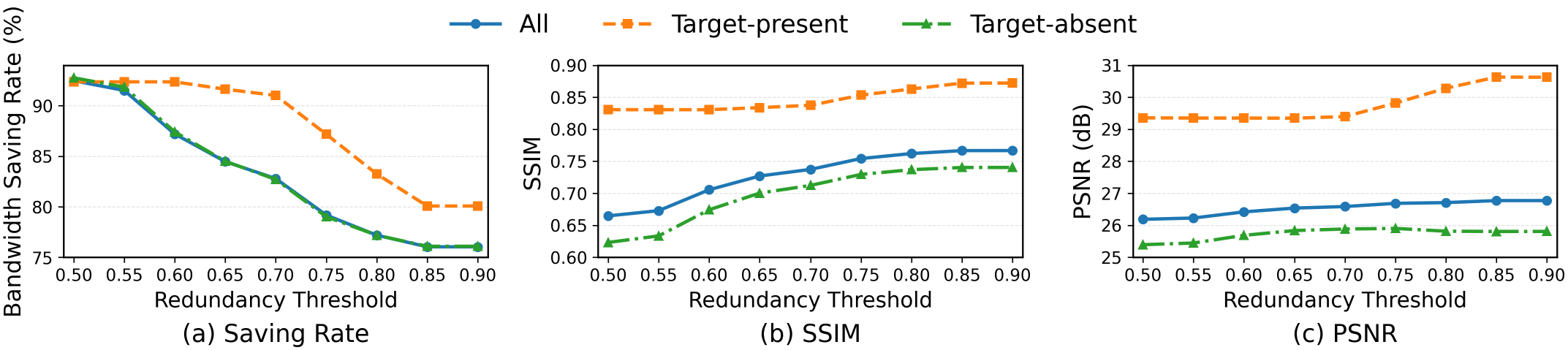}
    	\caption{Impact of the redundancy threshold on data reduction ratio, SSIM, and PSNR.}
    	\label{fig:threshold_sweep}
    \end{figure*} 
    
    To support content-adaptive compression, the raw HR image is first divided into non-overlapping $256\times256$ pixel tiles. SpaceRipple then estimates the redundancy of each tile before transmission. For a tile $I$ and a candidate downsampling factor $K$, the redundancy score is defined as
    \begin{equation}
    	\phi(I,K)=\operatorname{SSIM}\bigl(\operatorname{LI}(S(I,K)), I\bigr),
    	\label{eq:redundancy_phi}
    \end{equation}
    where $S(I,K)$ denotes downsampling $I$ by a factor of $K$, and $\operatorname{LI}(\cdot)$ denotes Lanczos interpolation back to the original tile size. A larger $\phi(I,K)$ indicates that the tile remains structurally recoverable after downsampling and can therefore tolerate stronger compression.
    
    For each tile, the sensing satellite selects the largest feasible $K$ satisfying
    \begin{equation}
    	\phi(I,K) > \tau_{\phi},
    	\label{eq:redundancy_threshold}
    \end{equation}
    where $\tau_{\phi}$ is the redundancy threshold. This rule enables aggressive compression for structurally redundant regions while retaining sufficient recoverable information for downstream semantic inference.
    
    After receiving the compressed representation and metadata, the edge computing satellite performs decompression-based restoration and lightweight preprocessing. Since compression may weaken textures, edges, and color consistency, this stage aims to recover a feature-compatible input for semantic inference rather than a visually perfect reconstruction \cite{guo2024kbisc}. The restoration process in \eqref{eq:restore} is conditioned on both the compressed image and metadata, allowing compression-ratio and calibration cues to stabilize recovery before semantic inference.

    \subsection{Compression-Aware MoE Enhancement}

    To improve semantic robustness after compression, SpaceRipple employs an MoE enhancement module on the edge computing satellite \cite{riquelme2021vmoe,sun2025deepspace}. In the proposed framework, MoE is used as a conditional enhancement mechanism rather than as a standalone architectural contribution. The routing decision is guided jointly by two types of cues: local pixel-feature characteristics extracted from different spatial regions of the compressed image, and the compression ratio contained in the metadata. The former provides content-aware cues for distinguishing regions with different structural and semantic importance, whereas the latter provides compression-aware cues indicating the severity of transmission-induced degradation.

    By combining these cues, the MoE module adaptively allocates enhancement strength to different regions and produces features that are more suitable for downstream semantic inference. This design does not rely on a particular number of experts or a specific routing implementation. Instead, the MoE module is used to preserve task-relevant information under high compression ratios and to support more aggressive yet reliable compression.
    
    \subsection{Semantic Output and Downlink Strategy}

    The enhanced features are then fed into the semantic head to generate mission outputs such as scene labels, detection results, regional descriptions, or event indicators. Unlike conventional image-downlink pipelines, SpaceRipple treats semantic results as the primary downlink payload, reserving link resources for high-value mission information and reducing response latency \cite{xin2024semsurvey,guo2024kbisc,dong2024lowbit}.

    In this way, the framework shifts the downlink objective from pixel-level delivery to mission-oriented information delivery. Compression, restoration, and enhancement are therefore not treated as isolated processing stages, but as coordinated components serving reliable semantic delivery under constrained inter-satellite communication resources.

    \section{Experimental Results and Analysis}

    \subsection{Evaluation Protocol}

    We evaluate SpaceRipple from three aspects: semantic-task performance, perceptual consistency after on-board restoration, and transmission efficiency. Let precision and recall be denoted by $P$ and $R$, respectively. The F1 score is
    \begin{equation}
        F1 = \frac{2PR}{P+R}.
    \end{equation}
    A higher F1 score indicates better retention of mission-relevant discriminative information after compression and on-board processing.

    For perceptual assessment, we adopt LPIPS \cite{zhang2018lpips}:
    \begin{equation}
        \text{LPIPS}(I_R,I_{HR})=
        \sum_{l} w_l
        \left\|
        \psi_l(I_R)-\psi_l(I_{HR})
        \right\|_2^2,
    \end{equation}
    where $\psi_l(\cdot)$ denotes deep features at layer $l$ and $w_l$ is the corresponding weight. In addition to task and perceptual metrics, we report model scale, contact-window throughput, and data reduction ratio. All methods are evaluated under the same testing conditions for fair comparison.

    Fig.~\ref{fig:threshold_sweep} shows the effect of the redundancy threshold on bandwidth saving and reconstruction quality. As the threshold increases, the data reduction ratio generally decreases, especially for target-containing regions, while PSNR and SSIM improve or remain stable. This indicates a clear tradeoff: a higher threshold preserves more task-relevant visual information, but at the cost of reduced transmission savings.

    \subsection{Lightweight Deployment and Task Performance}

    SpaceRipple adopts a two-tier deployment strategy. The sensing satellite executes lightweight compression and metadata generation, while the edge computing satellite performs restoration and semantic inference \cite{wang2025access,chen2024spaceedge}. The lightweight claim in this work primarily concerns deployment on the sensing satellite. In the current implementation, the auxiliary compression model on the sensing node contains only 1.53M parameters, while more computation-intensive modules, including restoration, MoE enhancement, and semantic inference, are shifted to the edge computing satellite.

    To further formalize this deployment characteristic, we define the effective sensing-side model footprint as
    \begin{equation}
        P_{\text{deploy}} = P_{\text{obs}},
    \end{equation}
    where $P_{\text{obs}}$ denotes the parameter count of the model actually deployed on the sensing satellite. In the current implementation, $P_{\text{obs}}=1.53$M, which indicates that the sensing-side model footprint remains low.

    The system-level computation distribution can be written as
    \begin{equation}
        \mathcal{C}_{\text{sys}}=\mathcal{C}_{\text{obs}}+\mathcal{C}_{\text{cs}},
    \end{equation}
    where $\mathcal{C}_{\text{obs}}$ and $\mathcal{C}_{\text{cs}}$ denote the computation assigned to the sensing satellite and the edge computing satellite, respectively. SpaceRipple is designed such that $\mathcal{C}_{\text{obs}} \ll \mathcal{C}_{\text{cs}}$.

\begin{figure*}[!t]
    	\centering
    	\includegraphics[width=0.88\textwidth]{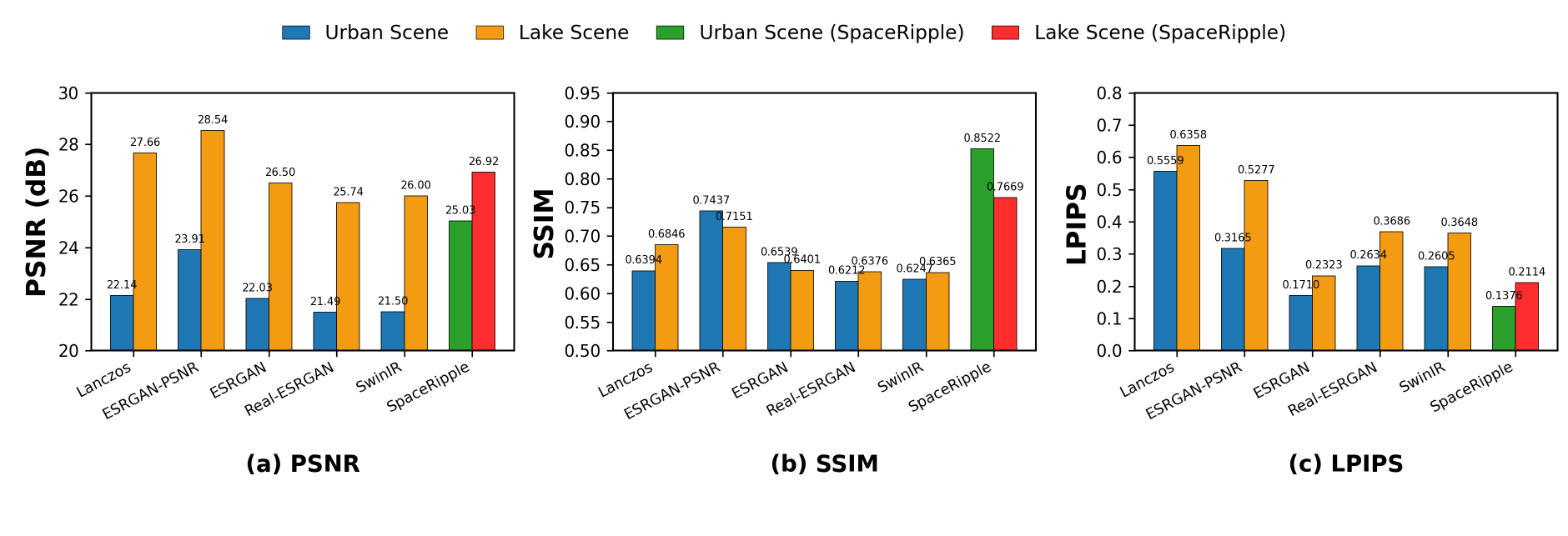}
    	\caption{Comparison of reconstruction quality with representative methods in terms of PSNR, SSIM, and LPIPS.}
    	\label{fig:MethodCompare}
    \end{figure*}

    Fig.~\ref{fig:MethodCompare} compares the reconstruction quality of SpaceRipple with representative methods in terms of PSNR, SSIM, and LPIPS. SpaceRipple achieves the best values on all three metrics. In particular, it reaches the lowest LPIPS of 0.2114 and the highest SSIM of 0.7689, indicating better perceptual similarity and structural preservation. The PSNR result of 26.92 dB is likewise the highest among the compared methods. Overall, the proposed restoration pipeline maintains favorable image quality while providing a stronger basis for downstream semantic inference.

    To complement the quantitative results, we present a qualitative example of the compression--reconstruction process.

    \begin{figure}[!t]
        \centering
        \includegraphics[width=0.88\columnwidth]{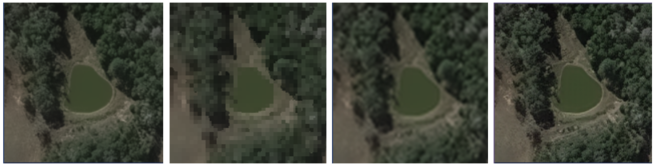}
        \caption{Qualitative comparison of the compression--reconstruction process. From left to right: original HR image, compressed SR image, upsampled SR image, and reconstructed HR image.}
        \label{fig:qualitative_reconstruction}
    \end{figure}

    As shown in Fig.~\ref{fig:qualitative_reconstruction}, the compressed image preserves the overall scene structure but loses fine textures. Upsampling restores more complete global content, although local regions remain blurred. The reconstructed HR image recovers sharper boundaries and more natural textures, consistent with the quantitative gains reported in Fig.~\ref{fig:MethodCompare}.

    \begin{table}[!t]
    	\centering
    	\vspace{0.06in}
    	        \caption{Comparison of target-presence detection performance among different methods on ship recognition and city vehicle recognition tasks.}
        \label{tab:detection_two_tasks}

        \begin{subtable}[t]{\linewidth}
            \centering
            \caption{Ship recognition task}
            \label{tab:ship_detection_metrics}
            \begin{tabular}{lccc}
                \toprule
                Method & Precision & Recall & F1 Score \\
                \midrule
                Bicubic     & 0.5870 & 0.7714 & 0.6667 \\
                Lanczos     & 0.5952 & 0.7142 & 0.6494 \\
                SwinIR      & 0.7727 & 0.9714 & 0.8608 \\
                Real-ESRGAN & 0.6957 & 0.9143 & 0.7901 \\
                SpaceRipple & 0.8421 & 0.8860 & 0.8635 \\
                \bottomrule
            \end{tabular}
        \end{subtable}

        \vspace{0.8em}

        \begin{subtable}[t]{\linewidth}
            \centering
            \caption{City vehicle recognition task}
            \label{tab:city_vehicle_detection_metrics}
            \begin{tabular}{lccc}
                \toprule
                Method & Precision & Recall & F1 Score \\
                \midrule
                Bicubic     & 0.8133 & 0.6224 & 0.7052 \\
                Lanczos     & 0.7568 & 0.5714 & 0.6512 \\
                SwinIR      & 0.6190 & 0.9286 & 0.7429 \\
                Real-ESRGAN & 0.6233 & 0.9286 & 0.7459 \\
                SpaceRipple & 0.8350 & 0.9290 & 0.8790 \\
                \bottomrule
            \end{tabular}
        \end{subtable}
    \end{table}

    As shown in Table~\ref{tab:detection_two_tasks}, SpaceRipple achieves the highest F1 score on both the ship recognition task and the city vehicle recognition task, demonstrating its superior and consistent target-presence detection performance across different scenarios.

    \subsection{Contact-Window Throughput Gain}

    To measure communication efficiency, let $W$ be the available contact window, 
    $C$ the link bandwidth, and $B_{\text{pkt}}$ the average transmitted payload 
    per sample. This formulation follows the practical observation that Earth-observation LEO satellites face intermittent and capacity-limited contact opportunities for space-to-ground data transfer \cite{sun2025deepspace}. The number of deliverable samples can be written as
    \begin{equation}
    	N = \left\lfloor \frac{CW}{B_{\text{pkt}}} \right\rfloor .
    \end{equation}
    A smaller payload therefore directly increases the number of observations that 
    can be delivered within the same contact window. This improvement is a 
    system-level gain rather than a codec-level gain alone. Instead of spending 
    most of the transmission budget on redundant pixel-level delivery, SpaceRipple 
    allocates more resources to semantically valuable observations, leading to 
    higher effective sensing throughput under bandwidth-constrained conditions.

    \begin{figure}[!t]
    	\centering
    	\includegraphics[width=0.88\columnwidth]{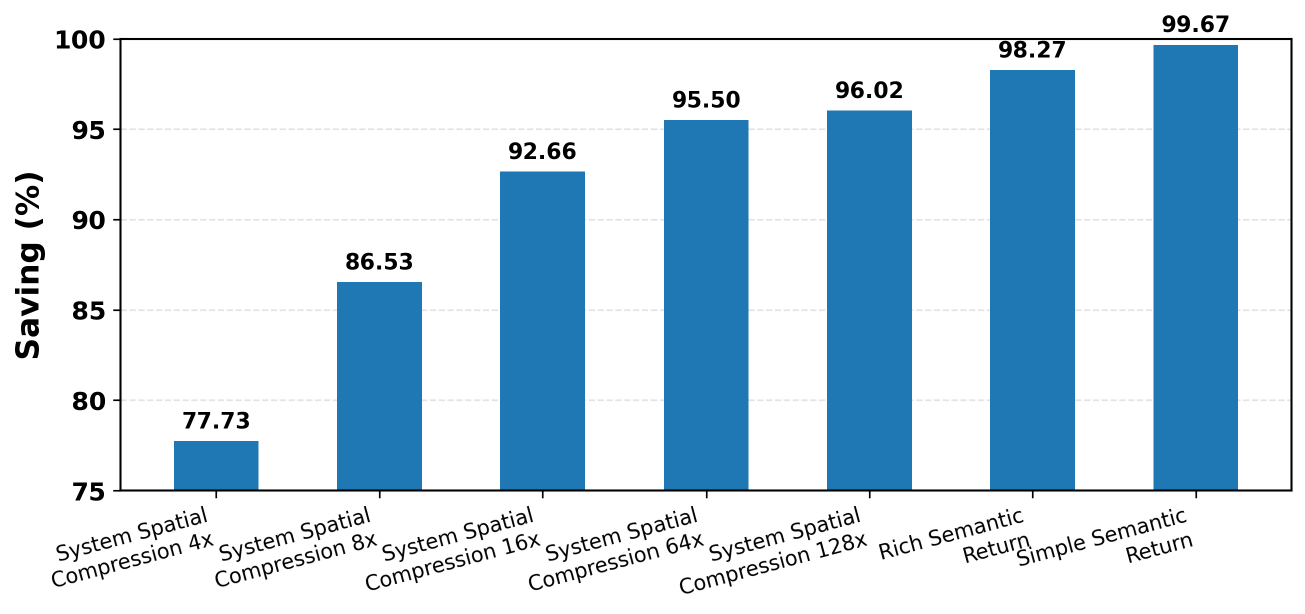}
    	\caption{Data reduction and semantic delivery accuracy for different downlink granularities.}
    	\label{fig:SemanticCompareBaseline}
    \end{figure}

    \subsection{Data Reduction and Latency Benefits of Semantic Delivery}

    For different downlink granularities, the total data reduction ratio is defined as
    \begin{equation}
        \eta = 1-\frac{B_{\text{SR}}}{B_{\text{base}}},
    \end{equation}
    where $B_{\text{SR}}$ is the total transmitted data under SpaceRipple and $B_{\text{base}}$ is that of the image-level downlink baseline.

    When only compact semantic results are downlinked, the transmitted data volume can be substantially reduced compared with image-level downlink~\cite{xin2024semsurvey,guo2024kbisc}. This is especially attractive for event-triggered monitoring and alerting, where a semantic alert or regional status description is often more valuable than delayed delivery of the full image.

    Fig.~\ref{fig:SemanticCompareBaseline} compares the data reduction ratios of different downlink strategies. 
    For image-oriented downlink, increasing the spatial compression ratio from 4$\times$ to 128$\times$ improves the data reduction ratio from 77.73\% to 96.02\%, but stronger compression also increases the reconstruction burden for preserving task-relevant information. Semantic-oriented delivery achieves higher efficiency, with data reduction ratios of 98.27\% for rich semantic delivery and 99.67\% for simple semantic delivery. Here, simple semantic delivery is encoded as a minimal JSON semantic packet, while rich semantic delivery is modeled as a structured detection message containing bounding boxes, class labels, confidence scores, and an image-level header. The default rich semantic payload is set to 192~B per target-containing image, with sensitivity evaluated from 128 to 512~B. These results show that semantic delivery reduces bandwidth more effectively than image-oriented downlink while avoiding costly reconstruction, making it better suited for mission-oriented Earth observation under limited link resources.

    Recent semantic image communication results below 0.03 bits per pixel further indicate that savings above 90\%, and in some cases above 99\%, are feasible when the delivery target is semantic rather than visual fidelity~\cite{dong2024lowbit}. This is consistent with the design of SpaceRipple.

    \subsection{System-Level Interpretation}

    The advantage of SpaceRipple appears consistently at the visual, task, and system levels. The pipeline maintains favorable PSNR, SSIM, and LPIPS after compression and recovery; task-level results confirm stronger preservation of downstream semantic information; and the contact-window throughput gain shows that these improvements also increase the number of mission-effective samples processed within a constrained communication opportunity. This interpretation is consistent with the design relation in \eqref{eq:jsys}, where task effectiveness is favored while payload size and the sensing-side model footprint are suppressed.

    \subsection{Discussion}

    The main value of SpaceRipple lies not only in stronger compression, but also in redefining the final downlink objective. Unlike conventional pipelines that center on image downlink and defer semantic processing to the ground, the proposed framework treats semantic information as the primary mission output and performs restoration only to the extent needed for reliable downstream inference.

    The framework also separates sensing and computing responsibilities in a deployment-friendly way. The sensing satellite carries only a lightweight front-end model, while restoration and semantic processing are shifted to the edge computing satellite. In the current implementation, this auxiliary compression model contains only 1.53M parameters, reducing the on-board model footprint at the sensing satellite.

    \section{Conclusion}

    This paper presented SpaceRipple, a lightweight framework for mission-oriented semantic delivery and on-board processing in Earth observation satellite networks. By coordinating sensing-side compression, edge-side restoration, and semantic-oriented delivery within a unified pipeline, SpaceRipple shifts the downlink objective from full-image transmission to mission-relevant information delivery under constrained satellite-network resources.

    Experimental results show that the proposed framework maintains competitive visual recoverability and effective task performance, while substantially improving contact-window throughput. These results indicate that SpaceRipple provides a practical way to support mission-oriented Earth observation when communication resources are limited.

    Future work will consider real link conditions, multi-task scenarios, and lighter deployment strategies.

\end{document}